\pdfoutput=1

\documentclass[11pt]{article}
\usepackage{xcolor}
\usepackage{colortbl}

\usepackage[]{acl}
\usepackage{amsfonts}
\usepackage{amssymb}
\usepackage{pifont}
\usepackage{tabularx} 
\usepackage{arydshln}

\usepackage{mathtools}

\usepackage{adjustbox}
\usepackage{multirow}
\usepackage{paralist}
\usepackage{CJKutf8}
\usepackage{booktabs}
\usepackage{setspace}
\usepackage{enumerate}
\usepackage{bbding} 
\usepackage{wasysym}
\usepackage{caption}
\usepackage{subcaption}
\usepackage{bbold}
\usepackage{tikz}
\usepackage{graphicx}
\usepackage{enumitem}
\usepackage{url}
\usepackage{hyperref}

\usepackage{times}
\usepackage{latexsym}
\usepackage{tikz}
\usepackage[edges]{forest}
\usepackage{pgfplots}
\usepackage{custom}
\usepackage{amsmath}
\usepackage{bm}
\usepackage{scalefnt}
\usepackage[most]{tcolorbox}
\usepackage[edges]{forest}
\usepackage[framemethod=tikz]{mdframed}
\usepackage{subcaption}
\usepackage{soul}

\usepackage{multirow}
\usepackage[T1]{fontenc}

\usepackage[utf8]{inputenc}

\usepackage{microtype}

\usepackage{inconsolata}
\usepackage{courier}

\title{\includegraphics[width=30pt]{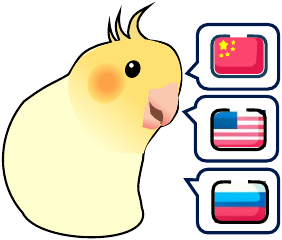}~Multilingual Large Language Model: \\A Survey of Resources, Taxonomy and Frontiers}
\definecolor{hidden-red}{RGB}{205, 44, 36}
\definecolor{hidden-blue}{RGB}{194,232,247}
\definecolor{hidden-orange}{RGB}{243,202,120}
\definecolor{hidden-green}{RGB}{34,139,34}
\definecolor{hidden-pink}{RGB}{255,245,247}
\definecolor{hidden-black}{RGB}{20,68,106}
\definecolor{purple}{RGB}{144,153,196}
\definecolor{yellow}{RGB}{255,228,123}
\definecolor{tkcolor}{RGB}{224,223,255}
\newtcolorbox{takeaways}[2][]{
	width=\columnwidth,
	colback = tkcolor, 
	colframe = tkcolor, 
	boxsep=0pt,left=10pt,right=10pt,top=0pt,bottom=0pt,
	fontupper=\linespread{0.9}\selectfont,
	title=#2,#1}

\newcommand{\eg}{\textit{e.g.,}}

\author{
	Libo Qin$^{\clubsuit}$\thanks{\ \ Equal Contribution} \quad Qiguang Chen$^{\spadesuit}$\footnotemark[1] \quad Yuhang Zhou$^{\spadesuit}$ \quad Zhi Chen$^{\diamondsuit}$ \quad Yinghui Li$^{\natural}$
	\\ 
	\textbf{Lizi Liao$^{\sharp}$ \quad Min Li$^{\clubsuit}$ \quad Wanxiang Che$^{\spadesuit}$ \quad Philip S. Yu$^{\heartsuit}$} \\
	\normalsize
	$^{\clubsuit}$ Central South University \quad
	$^{\spadesuit}$ Harbin Institute of Technology \quad $^{\diamondsuit}$ Shanghai AI Laboratory \\
	\normalsize
	$^{\natural}$ Tsinghua University \quad
	$^{\sharp}$ Singapore Management University \quad
	$^{\heartsuit}$ University of Illinons at Chicago \\
	\normalsize
	\texttt{lbqin@csu.edu.cn}, \texttt{\{qgchen,car\}@ir.hit.edu.cn}\\
}

\begin{document}
\maketitle
\begin{abstract} 
Multilingual Large Language Models are capable of using powerful Large Language Models to handle and respond to queries in multiple languages, which achieves remarkable success in multilingual natural language processing tasks.
Despite these breakthroughs, there still remains a lack of a comprehensive survey to summarize existing approaches and recent developments in this field. To this end, in this paper, we present a thorough review and provide
a unified perspective to summarize the recent progress as well as emerging trends in multilingual large language models (MLLMs) literature.
The contributions of this paper can be summarized:
(1) \textbf{\textit{First survey}}: to our knowledge, we take the first step and present a thorough review in MLLMs research field according to multi-lingual alignment;
(2) \textbf{\textit{New taxonomy}}: we offer a new and unified perspective to summarize the current progress of MLLMs;
(3) \textbf{\textit{New frontiers}}: we highlight several emerging frontiers and discuss the corresponding challenges; (4) \textbf{\textit{Abundant resources}}: we collect abundant open-source resources, including relevant papers, data corpora, and leaderboards.
We hope our work can provide the community with quick access and spur breakthrough research in MLLMs.

\end{abstract}
\section{Introduction}
In recent years, remarkable progress has been witnessed in large language models (LLMs)~\cite{brown2020language,touvron2023llama,bang2023multitask, zhao2023survey}, which have achieved excellent performance on various natural language processing tasks~\cite{pan2023preliminary,nguyen-etal-2023-cof,trivedi-etal-2023-interleaving}.
In addition, LLMs raise surprising emergent capabilities, including in-context learning~\cite{min2022rethinking,dong2022survey}, chain-of-thought reasoning~\cite{wei2022chain,huang2023not,qin2023cross}, and even planning~\cite{driess2023palm,hu2023tree}. 
Nevertheless, the majority of LLMs are English-centric, primarily focusing on English tasks~\cite{held2023material,zhang2023don}, which makes them somewhat weak for multilingual settings, especially in low-resource scenarios.
\begin{figure}[t]
	\centering
	\includegraphics[width=0.48\textwidth]{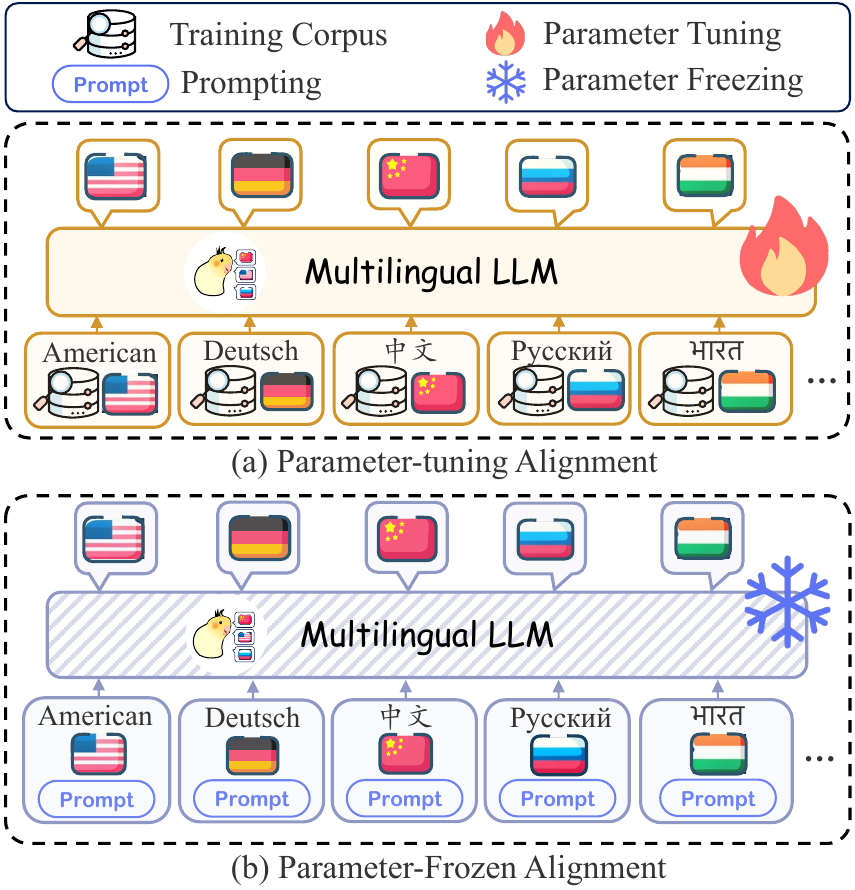}
	\caption{Parameter-Tuning Alignment ($\S$\ref{sec:parameter_tuned_align}) v.s. Parameter-Frozen Alignment ($\S$\ref{sec:param_freeze_align}). The former requires the model to fine-tune the MLLM parameters for cross-lingual alignment, while the latter directly uses prompts for alignment without parameter tuning.
	}
	\label{fig:intro}
\end{figure}
\begin{figure*}[t]
	\centering
	\includegraphics[width=0.98\textwidth]{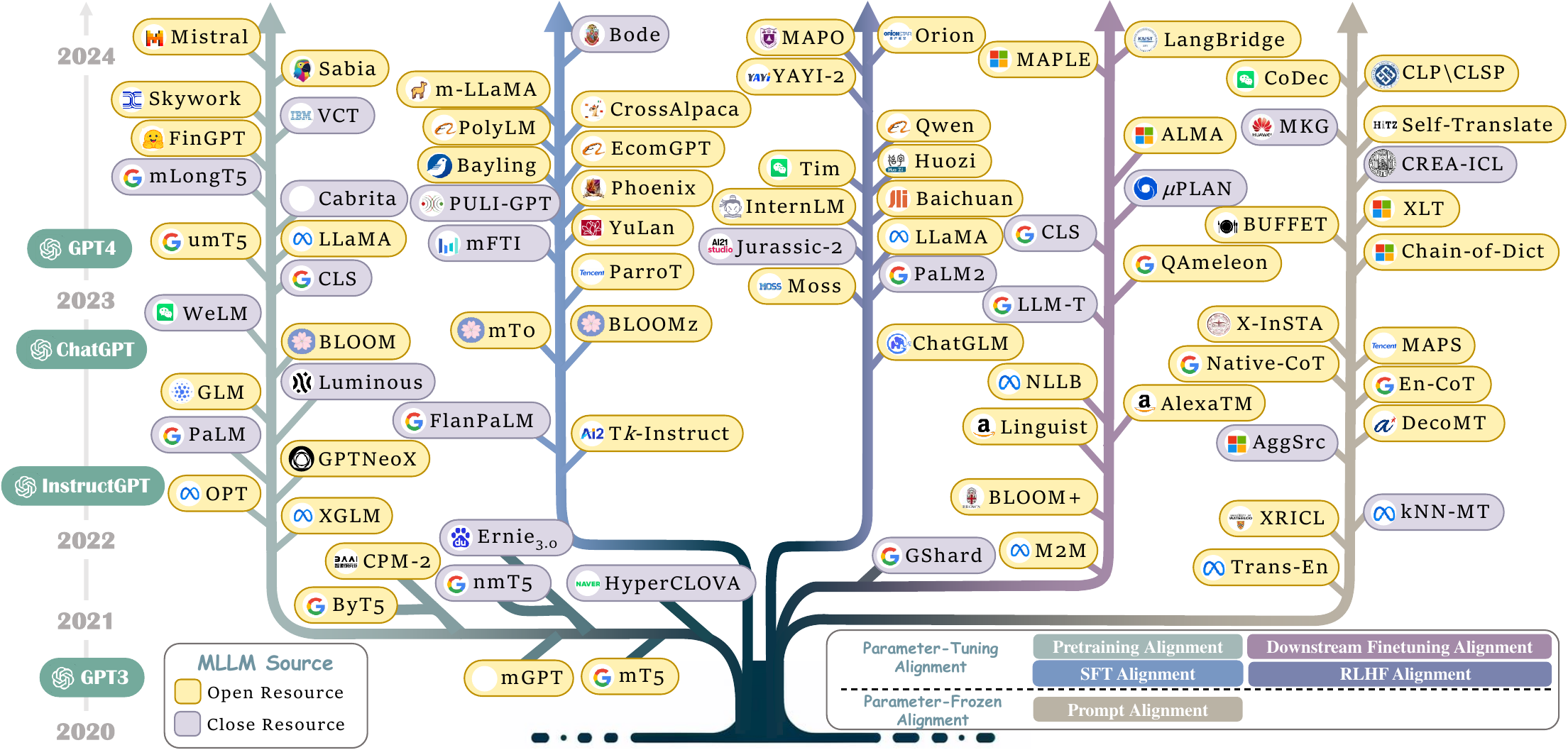}
	\caption{Evolution of selected MLLMs over the past five years, where colored branches indicate different alignment stages. For models with multiple alignment stages, the final stage is represented.}
	\label{fig:main}
\end{figure*}

Actually, there are over 7,000 languages in the world. With the acceleration of
globalization, the success of large language models should be considered to serve diverse countries and languages. To this end, multilingual large language models (MLLMs) possess the advantage of comprehensively handling multiple languages, gaining increasing attention.
Specifically, the existing MLLMs can be broadly divided into two groups based on different stages.
The first series of works~\cite{xue2020mt5,workshop2022bloom,zhang2023bayling,muennighoff2022crosslingual} leverage multilingual data to tuning the parameters to boost the overall multilingual performance. 
The second series of work~\cite{shi2022language,qin2023crosslingual,huang2023not} also adapt the advanced prompting strategies to unlock deeper multilingual potential of MLLMs during parameter-frozen inference stage.

While remarkable success has been achieved in the MLLMs, there still remains a lack of a comprehensive review and analysis of recent efforts in the literature, which hinders the development of MLLMs. To bridge this gap, we make the first attempt to conduct a comprehensive and detailed analysis of MLLMs.
Concretely, we first introduce the widely used data resource ($\S \ref{sec:data}$).  
Furthermore, due to the key challenge of alignment across languages, we introduce a novel taxonomy according to alignment strategies ($\S\ref{sec:taxonomy}$), aiming to provide a unified perspective in the literature, which includes:
\textbf{\textit{{parameter-tuning alignment}}} and \textbf{\textit{{parameter-frozen alignment}}} (as shown in Figure~\ref{fig:intro}).
Specifically, \textit{parameter-tuning alignment} requires the fine-tuning of model parameters to enhance alignment between English and target languages during pre-training, supervised fine-tuning, reinforcement learning from human feedback and downstream fine-tuning.
\textit{parameter-frozen alignment} refers to the  alignment achieved by prompting across languages that can be achieved without the need for parameter tuning.
Finally, we point out some potential frontier areas as well as the corresponding challenges for MLLMs, hoping to inspire the follow-up research ($\S \ref{sec:future-work-new-froniter}$).

The contributions of this work can be summarized as follows:
(1) \textbf{\textit{First survey}}: To the best of our knowledge, we are the first to present a comprehensive survey in the MLLMs literature according to multi-lingual alignment;
(2) \textbf{\textit{New taxonomy}}: We introduce a novel taxonomy categorizing MLLMs into two alignment types: \textit{parameter-frozen} and \textit{parameter-tuning}, offering a unified view for understanding the MLLMs literature;
(3) \textbf{\textit{New frontiers}}: We discuss some emerging frontiers and highlight their challenges as well as opportunities, hoping to pave the way for future research developments;
(4) \textbf{\textit{Exhaustive resources}}: We make the first attempt to organize MLLMs resources including open-source software, diverse corpora, and a curated list of relevant publications, accessible at \url{https://multilingual-llm.net}.

We hope that this work can serve as a valuable resource for researchers and inspire more breakthroughs in future research\footnote{ Figure~\ref{fig:main} illustrates the evolution of selected MLLMs over the past five years.}.

\section{Preliminary}

In this section, we will formally describe the definitions of monolingual large language model ($\S \ref{sec:mollm}$) and multilingual large language model ($\S \ref{sec:mllm}$).
\subsection{Monolingual Large Language Model}\label{sec:mollm}
\begin{figure}[t]
	\centering
	\includegraphics[width=0.48\textwidth]{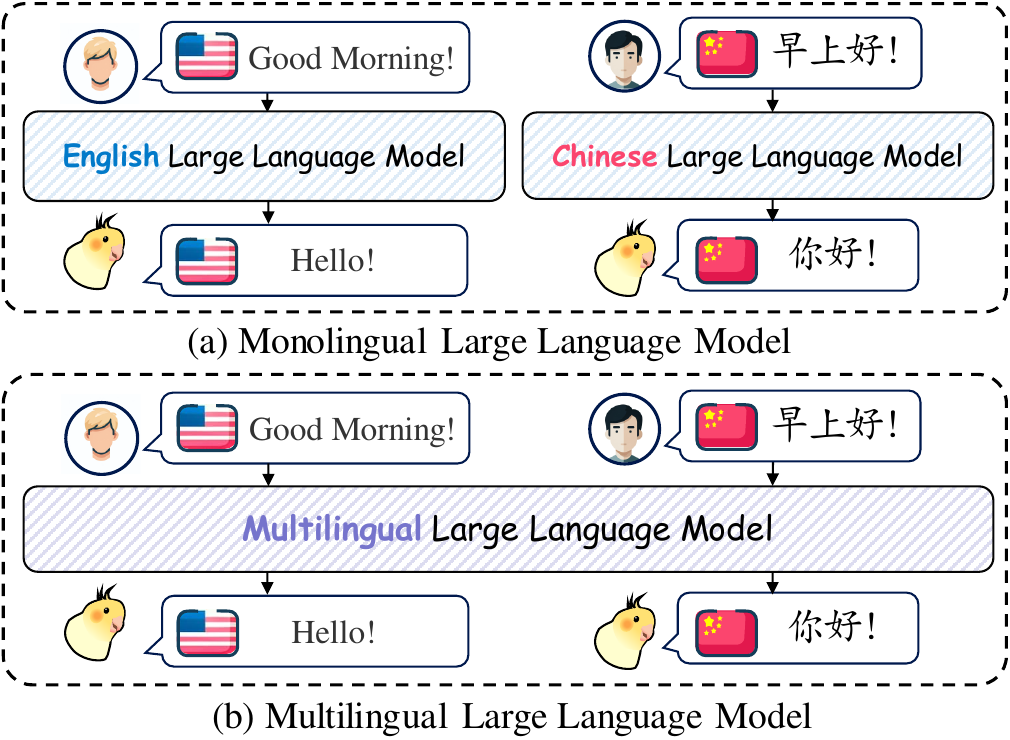}
	\caption{ Monolingual Large Language Model v.s.  Multilingual Large Language Model.
	}
	\label{fig:comparison}
\end{figure}

Monolingual large language models (LLM) can only process one language at a time. For example, as illustrated in Figure~\ref{fig:comparison} (a), English and Chinese LLM can separately handle English and Chinese language, respectively.
Formally, considering a set of languages $\mathcal{L} = \{\mathcal{L}_i\}^{|\mathcal{L}|}_{i=0}$, given input utterance $\mathcal{X}_i \in \mathcal{L}_i$ in languages $\mathcal{L}_i$, the process of monolingual LLM ($\mathcal{M}_{\texttt{mono}}$) generating the output $\mathcal{Y}_i$ can be defined as:
\begin{equation}
	\mathcal{Y}_i = \begin{cases} \mathcal{M}_{\texttt{mono}}(\mathcal{X}_i, \mathcal{L}_i), & \texttt{mono} = \mathcal{L}_i; \\ \texttt{Unexpect}, & \texttt{mono} \neq \mathcal{L}_i,
	\end{cases}
\end{equation}
where $\texttt{Unexpect}$ indicates that the LLM generates output in an unintended language; $\texttt{mono}$ denotes the single language.
\subsection{Multilingual Large Language Model}\label{sec:mllm}
As shown in Figure~\ref{fig:comparison} (b), unlike monolingual LLM, a multilingual LLM is capable of handling and producing content in various languages simultaneously, like English and Chinese. 
Formally, for MLLM $\mathcal{M}_{\texttt{multi}}$, where $\texttt{multi} \subseteq \mathcal{L}$ and $|\texttt{multi}| \geq 2$, the model's response is given by:
\begin{equation}
	\mathcal{Y} = 
		\mathcal{M}_{\texttt{multi}}(\mathcal{X}),
\end{equation}
where $\mathcal{X}$ and $ \mathcal{Y}$ belong to multiple languages, $\texttt{multi}$.

\section{Data Resource}\label{sec:data}

In this section, we describe the widely used data resources in pre-training ($\S \ref{sec:multilingual}$), supervised fine-tuning (SFT) ($\S \ref{sec:SFT}$) and reinforcement learning from human feedback (RLHF) ($\S \ref{sec:RLHF}$) stage~\citep{zhao2023survey} for multilingual large language model.
Detailed statistics can be found in Table~\ref{exp:pretrain-data} and Table~\ref{exp:sft-data} in the Appendix.

\subsection{Multilingual Pretraining Data}\label{sec:multilingual}
The widely used multilingual corpora for pre-training in MLLMs can be divided into 3 categories: (1) \textbf{\textit{Manual Creation}}: obtains high-quality pre-training corpora through manual creation and proofreading, which consists of the Bible Corpus~\cite{mayer-cysouw-2014-creating}  and MultiUN~\cite{ziemski-etal-2016-united}.
(2) \textbf{\textit{Web Crawling}}: involves crawling extensive multilingual data from the internet, which includes  OSCAR~\cite{suarez2019asynchronous}, CC-100~\cite{conneau2020unsupervised}, mC4~\cite{xue2021mt5} and Redpajama-v2~\cite{together2023redpajama}.
Another series of data are extracted from Wikipedia to enhance the knowledge of MLLMs. Common datasets include Wikipedia~\cite{wikidump} , WikiMatrix~\cite{schwenk-etal-2021-wikimatrix} and WikiExpl~\cite{han2023bridging}.
(3) \textbf{\textit{Benchmark Adaptation}}: 
means re-cleaning or integrating existing benchmarks to enhance data quality which includes OPUS-100~\cite{zhang-etal-2020-improving}, Culturax~\cite{nguyen2023culturax}, OPUS~\cite{tiedemann-2012-parallel}, WMT~\cite{kocmi2023findings} and ROOTS~\cite{laurenccon2022bigscience}.
\tikzstyle{my-box}=[
rectangle,
draw=hidden-black,
rounded corners,
text opacity=1,
minimum height=1.5em,
minimum width=5em,
inner sep=2pt,
align=center,
fill opacity=.5,
]
\tikzstyle{leaf}=[
my-box, 
minimum height=1.5em,
fill=yellow!32, 
text=black,
align=left,
font=\normalsize,
inner xsep=2pt,
inner ysep=4pt,
]
\tikzstyle{leaf2}=[
my-box, 
minimum height=1.5em,
fill=purple!27, 
text=black,
align=left,
font=\normalsize,
inner xsep=2pt,
inner ysep=4pt,
]

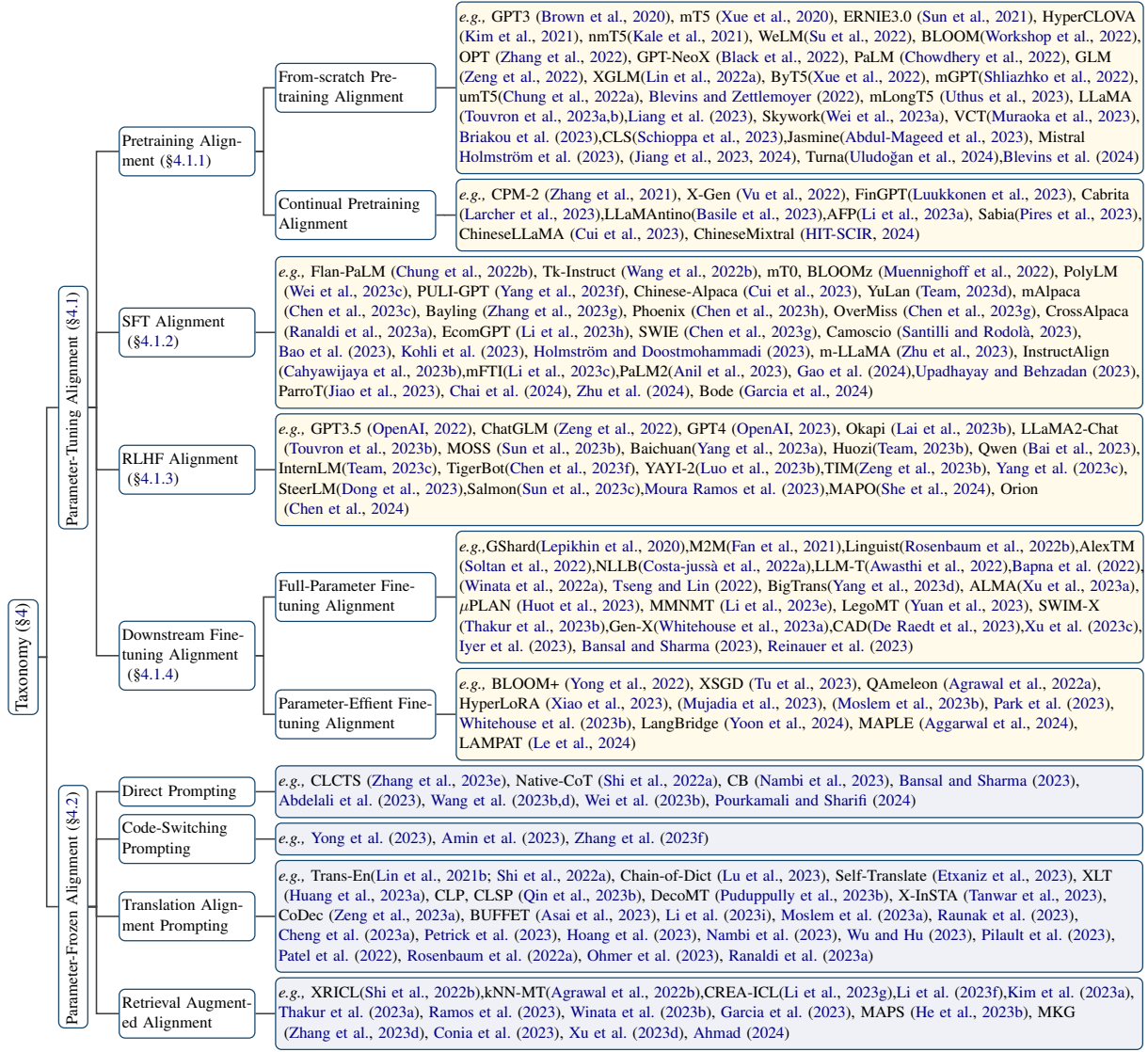
\begin{figure*}[t]
\vspace{-2mm}
\centering
\resizebox{\textwidth}{!}{
	\begin{forest}
		forked edges,
		for tree={
			grow=east,
			reversed=true,
			anchor=base west,
			parent anchor=east,
			child anchor=west,
			base=left,
			font=\large,
			rectangle,
			draw=hidden-black,
			rounded corners,
			align=left,
			minimum width=4em,
			edge+={darkgray, line width=1pt},
			s sep=3pt,
			inner xsep=2pt,
			inner ysep=3pt,
			line width=0.8pt,
			ver/.style={rotate=90, child anchor=north, parent anchor=south, anchor=center},
		},
		where level=1{text width=14.7em,font=\normalsize,}{},
		where level=2{text width=8.0em,font=\normalsize,}{},
		where level=3{text width=9.5em,font=\normalsize,}{},
		where level=4{text width=12em,font=\normalsize,}{},
		[Taxonomy~(\S\ref{sec:taxonomy}),ver
		[Parameter-Tuning  Alignment~(\S\ref{sec:parameter_tuned_align}),ver
			[Pretraining Align- \\ ment~(\S\ref{sec:pretrain_align})
				[From-scratch Pre-\\ training Alignment
					[\eg~GPT3~\cite{brown2020language}{,} mT5~\cite{xue2020mt5}{,}  ERNIE3.0~\cite{sun2021ernie}{,}  HyperCLOVA\\\cite{kim2021changes}{,}
					nmT5\cite{kale2021nmt5}{,} WeLM\cite{su2022welm}{,} BLOOM\cite{workshop2022bloom}{,}\\
					OPT~\cite{zhang2022opt}{,}  GPT-NeoX~\cite{black2022gpt}{,} 
					PaLM~\cite{chowdhery2022palm}{,} 
					GLM\\\cite{zeng2022glm}{,} XGLM\cite{lin-etal-2022-shot}{,}
					ByT5\cite{xue2022byt5}{,}
					mGPT\cite{shliazhko2022mgpt}{,} \\
					umT5\cite{chung2022unimax}{,}
					\citet{blevins2022language}{,}
					mLongT5~\cite{uthus2023mlongt5}{,} LLaMA\\\cite{touvron2023llama,touvron2023llama2}{,}\citet{liang2023multi}{,} Skywork\cite{wei2023skywork}{,} VCT\cite{muraoka2023cross}{,}\\\citet{briakou2023searching}{,}CLS\cite{schioppa2023cross}{,}Jasmine\cite{abdul2023jasmine}{,}  Mistral\\\citet{holmstrom2023bridging}{,} \cite{jiang2023mistral,jiang2024mixtral}{,} Turna\cite{uludougan2024turna}{,}\citet{blevins2024breaking}
					, leaf, text width=42em]
				]
				[Continual Pretraining \\ Alignment
					[\eg~CPM-2~\cite{zhang2021cpm}{,}  X-Gen~\cite{vu2022overcoming}{,}
					FinGPT\cite{luukkonen2023fingpt}{,} Cabrita\\\cite{larcher2023cabrita}{,}LLaMAntino\cite{basile2023llamantino}{,}AFP\cite{li2023align}{,}   Sabia\cite{pires2023sabi}{,}\\ ChineseLLaMA~\cite{cui2023efficient}{,} ChineseMixtral~\cite{Chinese-Mixtral}
					, leaf, text width=42em]
				]
			]
			[SFT Alignment\\~(\S\ref{sec:sft_align})
				[\eg~Flan-PaLM~\cite{chung2022scaling}{,} Tk-Instruct~\cite{wang2022super}{,} mT0{,} BLOOMz~\cite{muennighoff2022crosslingual}{,} PolyLM\\~\cite{wei2023polylm}{,} PULI-GPT~\cite{yang2023mono}{,}  Chinese-Alpaca~\cite{cui2023efficient}{,} YuLan~\cite{YuLan-Chat}{,} 
				mAlpaca \\~\cite{chen2023monolingual}{,} Bayling~\cite{zhang2023bayling}{,} Phoenix~\cite{chen2023phoenix}{,} OverMiss~\cite{chen2023improving}{,} CrossAlpaca\\~\cite{ranaldi2023empowering}{,} EcomGPT~\cite{li2023ecomgpt}{,} SWIE~\cite{chen2023improving}{,} Camoscio~\cite{santilli2023camoscio}{,} \\\citet{bao2023conversations}{,} \citet{kohli2023building}{,} \citet{holmstrom2023making}{,}  m-LLaMA~\cite{zhu2023extrapolating}{,} InstructAlign\\\cite{cahyawijaya2023instruct}{,}mFTI\cite{li2023eliciting}{,}PaLM2\cite{anil2023palm}{,} \citet{gao2024towards}{,}\citet{upadhayay2023taco}{,} \\ ParroT\cite{jiao2023parrot}{,} \citet{chai2024xcot}{,}  \citet{zhu2024question}{,}  Bode~\cite{garcia2024introducing}
				, leaf, text width=53.1em]
			]
			[RLHF Alignment\\~(\S\ref{sec:rlhf_align})
				[\eg~GPT3.5~\cite{openai2022chatgpt}{,} ChatGLM~\cite{zeng2022glm}{,} GPT4~\cite{openai2023gpt4}{,}  Okapi~\cite{lai2023okapi}{,} LLaMA2-Chat\\~\cite{touvron2023llama2}{,}  MOSS~\cite{sun2023moss}{,} Baichuan\cite{yang2023baichuan}{,} Huozi\cite{huozi}{,} Qwen~\cite{bai2023qwen}{,} \\ InternLM\cite{team2023internlm}{,} TigerBot\cite{chen2023tigerbot}{,} YAYI-2\cite{luo2023yayi}{,}TIM\cite{zeng2023tim}{,}
				\citet{yang2023direct}{,}\\SteerLM\cite{dong2023steerlm}{,}Salmon\cite{sun2023salmon}{,}\citet{moura2023aligning}{,}MAPO\cite{she2024mapo}{,}
				Orion\\~\cite{chen2024orion}
				, leaf, text width=53.1em]
			]
			[Downstream Fine-\\tuning Alignment\\~(\S\ref{sec:ft_align})
				[Full-Parameter Fine-\\tuning Alignment
					[\eg GShard\cite{lepikhin2020gshard}{,}M2M\cite{fan2021beyond}{,}Linguist\cite{rosenbaum2022linguist}{,}AlexTM\\\cite{soltan2022alexatm}{,}NLLB\cite{costa2022no}{,}LLM-T\cite{awasthi2022bootstrapping}{,}\citet{bapna2022building}{,}\\
					\cite{winata-etal-2022-cross}{,} \citet{tseng2022enhancing}{,}   BigTrans\cite{yang2023bigtrans}{,}
					ALMA\cite{xu2023paradigm}{,}\\
					 $\mu$PLAN~\cite{huot2023mu}{,} MMNMT~\cite{li2023mmnmt}{,} LegoMT~\cite{yuan2023lego}{,}
					SWIM-X\\\cite{thakur2023leveraging}{,}Gen-X\cite{whitehouse2023llm}{,}CAD\cite{de2023zero}{,}\citet{xu2023structural}{,}\\
					\citet{iyer2023towards}{,}
					\citet{bansal2023large}{,}
					\citet{reinauer2023neural}
					, leaf, text width=42em]
				]
				[Parameter-Effient Fine-\\tuning Alignment
					[\eg~BLOOM+~\cite{yong2022bloom+}{,}   XSGD~\cite{tu2023efficiently}{,} QAmeleon~\cite{agrawal2022qameleon}{,}\\  HyperLoRA~\cite{xiao2023task}{,}
					 \cite{mujadia2023assessing}{,}
					 \cite{moslem2023fine}{,} \citet{park2023analysis}{,}\\ \citet{whitehouse2023parameter}{,}
					LangBridge~\cite{yoon2024langbridge}{,}
					MAPLE~\cite{aggarwal2024maple}{,}\\ LAMPAT~\cite{le2024lampat}
					, leaf, text width=42em]
				]
			]]
		[Parameter-Frozen Alignment~(\S\ref{sec:param_freeze_align}),ver
		[Direct Prompting
		[\eg~CLCTS~\cite{zhang2023cross}{,}  Native-CoT~\cite{shi2022language}{,}
		CB~\cite{nambi2023breaking}{,} \citet{bansal2023large}{,}  \\ \citet{abdelali2023benchmarking}{,} \citet{wang2023cross,wang2023document}{,} \citet{wei2023zero}{,} \citet{pourkamali2024machine}
		, leaf2, text width=53.1em]
		]
		[Code-Switching \\Prompting
		[\eg~\citet{yong2023prompting}{,} \citet{amin2023marathi}{,} \citet{zhang2023multilingual}
		, leaf2, text width=53.1em]
		]
		[Translation Align-\\ment Prompting
		[\eg~Trans-En\cite{lin2021few,shi2022language}{,}
		Chain-of-Dict~\cite{lu2023chain}{,} 
		Self-Translate~\cite{etxaniz2023multilingual}{,} XLT\\~\cite{huang2023not}{,} CLP{,} CLSP~\cite{qin2023crosslingual}{,}
		DecoMT~\cite{puduppully2023decomt}{,} X-InSTA~\cite{tanwar2023multilingual}{,}
		\\CoDec~\cite{zeng2023improving}{,} BUFFET~\cite{asai2023buffet}{,} \citet{li2023bilingual}{,} \citet{moslem2023adaptive}{,} \citet{raunak2023leveraging}{,} \\ \citet{cheng2023scale}{,} \citet{petrick2023document}{,} \citet{hoang2023fly}{,} \citet{nambi2023breaking}{,} \citet{wu2023exploring}{,} \citet{pilault2023interactive}{,} \\ \citet{patel2022bidirectional}{,} \citet{rosenbaum2022clasp}{,} \citet{ohmer2023evaluating}{,} \citet{ranaldi2023empowering}
		, leaf2, text width=53.1em]
		]
		[Retrieval Augment-\\ed Alignment
		[\eg~XRICL\cite{shi2022xricl}{,}kNN-MT\cite{agrawal2022context}{,}CREA-ICL\cite{li2023classification}{,}\citet{li2023crosslingual}{,}\citet{kim2023boosting}{,} \\
		 \citet{thakur2023nomiracl}{,}  \citet{ramos2023lmcap}{,}  \citet{winata2023multilingual}{,} \citet{garcia2023unreasonable}{,} 
		 MAPS~\cite{he2023exploring}{,} MKG\\~\cite{zhang2023leveraging}{,}
		\citet{conia2023increasing}{,} \citet{xu2023language}{,} \citet{ahmad2024enhancing}
		, leaf2, text width=53.1em]
		]
		]
		]
	\end{forest}
}
\vspace{-4mm}
\caption{Taxonomy of MLLMs which includes \textit{Parameter-Tuning Alignment Methodology} and \textit{Parameter-Frozen Alignment Methodology}.}
\label{fig:taxonomy}
\vspace{-3mm}
\end{figure*}
\subsection{Multilingual SFT Data}\label{sec:SFT}
Similarly, we categorize the existing multilingual SFT data into 4 classes:
(1) \textbf{\textit{Manual Creation}}: acquires SFT corpora through manual creation and proofreading, which includes Sup-NatInst~\cite{wang2022super}, OpenAssist~\cite{kopf2023openassistant} and COIG-PC$_\text{lite}$~\cite{COIG-PC}.
(2) \textbf{\textit{Machine Translation}}: translates the existing monolingual datasets into multilingual instruction datasets, which comprises xP3-MT~\cite{muennighoff2022crosslingual},  MGSM8K$_\text{Instruct}$~\cite{chen2023breaking}, CrossAlpaca~\cite{ranaldi2023empoweringt,cui2023efficient}, MultilingualSIFT~\cite{Chen_MultilingualSIFT_Multilingual_Supervised_2023} and Bactrain-X~\cite{li2023bactrian}.
(3) \textbf{\textit{Benchmark Adaptation}}: involves  transformation from  existing benchmarks to instruction format. Widely used datasets include xP3~\cite{muennighoff2022crosslingual}, PolyglotPrompt~\cite{fu2022polyglot}, and BUFFET~\cite{asai2023buffet}.
(4) \textbf{\textit{MLLMs Aided Generation}}: means that the data are automatically synthesized by the MLLMs, containing Vicuna~\cite{vicuna2023}, OverMiss~\cite{chen2023improving}, ShareGPT~\cite{sharegpt}, BELLE~\cite{belle2023exploring}, MultiAlpaca~\cite{wei2023polylm}, Guanaco~\cite{dettmers2023qlora} and Alpaca-4~\cite{peng2023instruction}.

\subsection{Multilingual RLHF Data}\label{sec:RLHF}
Some work leveraged the multilingual RLHF data to improve alignment.
Specifically, \citet{lai2023okapi} leverages multilingual ranking data for training a reward model using RLHF.
\citet{zeng2023tim} introduce the TIM dataset to train a more effective reward model in multilingual contexts.

\section{Taxonomy}\label{sec:taxonomy}
\label{sec:lang_align_method}
As shown in Figure~\ref{fig:taxonomy}, we introduce a novel taxonomy including \textit{parameter-tuning alignment} ($\S \ref{sec:parameter_tuned_align}$) and \textit{parameter-frozen alignment} ($\S \ref{sec:param_freeze_align}$), which aims to provide a unified view for researchers to understand the MLLMs literature.	
Specifically, parameter tuning alignment (PTA) comprises a series of progressively advanced training and alignment strategies, including  Pretraining Alignment, Supervised Fine-Tuning (SFT) Alignment, Reinforcement Learning from Human Feedback (RLHF) Alignment, and, ultimately, Downstream Fine-Tuning Alignment. These stages collectively aim to refine model parameters to align the multilingual performance systematically.
Conversely, the parameter frozen alignment (PFA) focuses on four prompting strategies based on PTA: Direct Prompting, Code-Switching Prompting, Translation Alignment Prompting, and Retrieval-Augmented Alignment. This method maintains the original model parameters to achieve desired outcomes.

\begin{figure*}[t]
	\centering
	\includegraphics[width=0.98\textwidth]{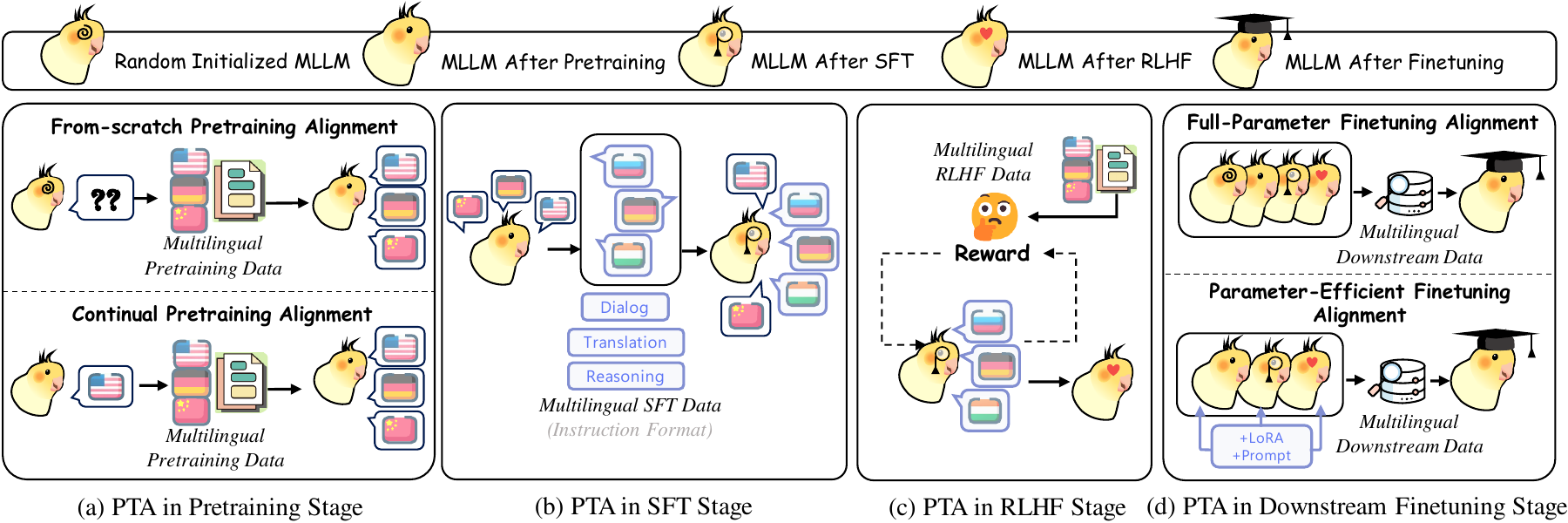}
	\caption{Overview of Parameter-Tuning Alignment ($\S$ \ref{sec:parameter_tuned_align}) Methods, which including \textit{PTA in Pretraining Stage} ($\S$ \ref{sec:pretrain_align}), \textit{PTA in SFT stage} ($\S$ \ref{sec:sft_align}), \textit{PTA in RLHF stage} ($\S$ \ref{sec:rlhf_align}) and \textit{PTA in Downstream Finetuning stage} ($\S$ \ref{sec:ft_align}).}
	\label{fig:framework}
\end{figure*}
\subsection{Parameter-Tuning Alignment}
\label{sec:parameter_tuned_align}
Parameter-tuning alignment indicates that MLLMs should tune their parameters for better cross-lingual alignment~\cite{wen2023hyperpolyglot}. 
As shown in Figure \ref{fig:framework}, we discuss the four categories of parameter-tuning alignment (PTA), including PTA in pretraining stage ($\S \ref{sec:pretrain_align}$), PTA in SFT stage ($\S \ref{sec:sft_align}$), PTA in RLHF stage ($\S \ref{sec:rlhf_align}$) and PTA in Finetuning stage ($\S \ref{sec:ft_align}$).

\subsubsection{PTA in Pretraining Stage}
\label{sec:pretrain_align}
\paragraph{\textit{From-scratch Pretraining Alignment}.}
A series of approaches have achieved to alignment across languages by  tuning the initially random parameters of MLLMs during pretraining (see Figure~\ref{fig:framework} (a)).
Specifically, ~\citet{blevins2022language,briakou2023searching,holmstrom2023bridging} observed that adding a few multilingual data during the from-scratch pretraining alignment, even unintentionally, can significantly boost the multilingual performance.
Inspired by this, \citet{zeng2022glm,su2022welm} used bilingual data in their from-scratch pretraining for alignment.
mT5~\cite{xue2020mt5}, Ernie3.0~\cite{sun2021ernie}, ByT5~\cite{xue2022byt5}, BLOOM~\cite{workshop2022bloom}, 
LLaMA~\cite{touvron2023llama2,touvron2023llama}, PaLM~\cite{chowdhery2022palm}, 
Mistral~\cite{jiang2023mistral}, Mixtral~\cite{jiang2024mixtral}, PolyLM~\cite{wei2023polylm}, 
\citet{kale2021nmt5,kim2021changes,shliazhko2022mgpt,chai2022ernie,schioppa2023cross,abdul2023jasmine,uthus2023mlongt5,wei2023skywork,uludougan2024turna}
incorporated multilingual data in pretraining stage for better  alignment.
\citet{blevins2024breaking} utilizes Mixture-of-Experts (MoE) to independently train language models on subsets of multilingual corpora to alleviate the problem of multilingual parameter competition.
Furthermore, to enhance the performance of low-resource languages, umT5~\cite{chung2022unimax} and XGLM~\cite{lin-etal-2022-shot} adopted equitable data sampling methods during from-scratch pretraining. \citet{muraoka2023cross} introduced VCT to leverage vision for indirect cross-lingual alignment in from-scratch pretraining.

\paragraph{\textit{Continual Pretraining Alignment.}}
To address the high computational cost of from-scratch pretraining, continual pretraining alignment builds the pretraining process upon pretrained MLLMs (as shown in Figure~\ref{fig:framework} (a)). Specifically, CPM-2~\cite{zhang2021cpm}, Sabia~\cite{pires2023sabi}, FinGPT~\cite{luukkonen2023fingpt}, X-Gen~\cite{vu2022overcoming}, AFP~\cite{li2023align}, Cabrita~\cite{larcher2023cabrita}, LLaMAntino~\cite{basile2023llamantino} focused on adding more target language data during continual pretraining for general performance.
Further,  \citet{cui2023efficient,Chinese-Mixtral} emphasized extending the MLLMs' vocabularies to adapt to new languages.

\subsubsection{PTA in SFT Stage}
\label{sec:sft_align}
As illustrated in Figure~\ref{fig:framework} (b), 
PTA in SFT stage means leveraging multiple multilingual task data with instruction format for tuning parameters~\cite{fu2022polyglot,yang2023mono,YuLan-Chat,chen2023monolingual,chen2023improving,ranaldi2023empowering,li2023ecomgpt,chen2023improving,santilli2023camoscio,bao2023conversations,kohli2023building,holmstrom2023making,garcia2024introducing}.
In particular, models like Flan-PaLM~\cite{chung2022scaling}, mT0, BLOOMz~\cite{muennighoff2022crosslingual}, PolyLM~\cite{wei2023polylm}, Tk-Instruct~\cite{wang2022super}, Chinese-Alpaca~\cite{cui2023efficient}, Bayling~\cite{zhang2023bayling} and Phoenix~\cite{chen2023phoenix}, 
directly incorporated multilingual data in the SFT stage to achieve implicit multilingual alignment across languages.
Besides, to solve the scarcity of multilingual SFT task data, PaLM2~\cite{anil2023palm}, \citet{zhu2023extrapolating,cahyawijaya2023instruct,li2023eliciting,gao2024towards} added translation task during the SFT alignment stage to improve alignment.
Further, \citet{upadhayay2023taco,chai2024xcot,zhu2024question} began to consider using a more effective SFT alignment strategy to optimize the reasoning process.

\begin{figure*}[t]
	\centering
	\includegraphics[width=\textwidth]{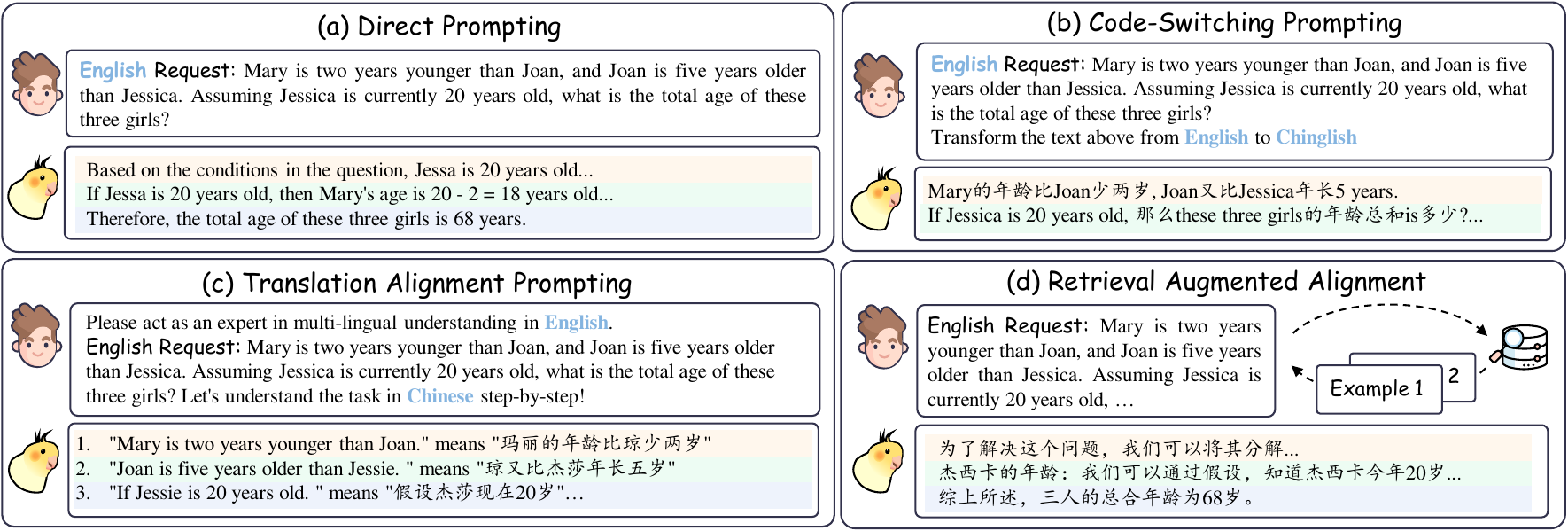}
	\caption{Overview of Parameter-Frozen Alignment ($\S$ \ref{sec:param_freeze_align}) methods, where prompts in sub-figures sourced from  \citet{qin2023crosslingual} and \citet{zhang2023multilingual}.}
	\label{fig:framework2}
\end{figure*}

\subsubsection{PTA in RLHF Stage}
\label{sec:rlhf_align}
As shown in Figure~\ref{fig:framework} (c), to achieve alignment in reinforcement learning from human feedback (RLHF) stage, Okapi~\cite{lai2023okapi}, LLaMA2-Chat~\cite{touvron2023llama2}, ChatGLM~\cite{zeng2022glm}, MOSS~\cite{sun2023moss}, Baichuan~\cite{yang2023baichuan}, Huozi~\cite{huozi}, Qwen~\cite{bai2023qwen},  InternLM~\cite{team2023internlm}, ParroT~\cite{jiao2023parrot}, TigerBot~\cite{chen2023tigerbot}, MOSS~\cite{sun2023moss}, YAYI-2~\cite{luo2023yayi}, \citet{yang2023direct,moura2023aligning} and Orion~\cite{chen2024orion} directly integrated multilingual RLHF data for training multilingual reward models.
Additionally, \citet{zeng2023tim,dong2023steerlm,she2024mapo} introduced a multilingual reward model to compare translation outputs across different granularity.
\citet{sun2023salmon} proposed a Salmon framework, to enhance multilingual RLHF by self-generating rewards for better alignment.

\subsubsection{PTA in Downstream Finetuning Stage}
\label{sec:ft_align}

\paragraph{\textit{Full-Parameter Finetuning Alignment}}
Full-parameter finetuning in MLLMs means tuning all parameters in downstream tasks (see Figure~\ref{fig:framework} (d)).
Specifically, GShard~\cite{lepikhin2020gshard}, Linguist~\cite{rosenbaum2022linguist}, \citet{fan2021beyond,bapna2022building,tseng2022enhancing,iyer2023towards}, NLLB~\cite{costa2022no} AlexTM~\cite{soltan2022alexatm}, and BigTrans~\cite{yang2023bigtrans} focused on directly fine-tuning the full parameters across various downstream tasks (e.g., information extraction, machine translation).
\citet{xu2023structural,huot2023mu,yuan2023lego,li2023mmnmt} proposed multi-step or fine-grained alignment strategies during full-parameter tuning.
Furthermore, to enhance the efficiency, \citet{awasthi2022bootstrapping,de2023zero,thakur2023leveraging,whitehouse2023llm,bansal2023large,xu2023paradigm,reinauer2023neural} focused on knowledge distillation from larger to smaller MLLMs.

\paragraph{\textit{Parameter-Efficient Finetuning Alignment}}
A series of studies employ Parameter-Efficient Finetuning (PEFT) alignment approaches for reducing full-parameter fine-tuning costs~\cite{yong2022bloom+,mujadia2023assessing,moslem2023fine}, which is shown in Figure~\ref{fig:framework} (d).
\citet{agrawal2022qameleon,tu2023efficiently,park2023analysis} proposed minimal soft prompt prefix fine-tuning for better alignment.
Furthermore, \citet{whitehouse2023parameter,xiao2023task,aggarwal2024maple,le2024lampat} proposed methods based on Low-Rank Adaptation (LoRA) to achieve PEFT alignment.
Further, \citet{yoon2024langbridge} introduced a LangBridge model to bridge multilingual encoder to single-lingual LLM to effectively achieve promising performance.

\begin{takeaways}
\ \paragraph{Takeaways} (1) \textit{PTA in pretraining stage brings the essential multilingual capabilities of the MLLMs.}
(2) \textit{The effectiveness of alignment in MLLMs is greatly influenced by previous alignment stage, (e.g. Pretraining will significantly influence SFT).}
\end{takeaways}

\subsection{Parameter-Frozen Alignment}
\label{sec:param_freeze_align}
In contrast to the traditional 
parameter-tuning approaches~\cite{zheng-etal-2022-hit}, parameter-frozen alignment methods aim to perform alignment without any parameter tuning.
\label{sec:const_align}
The most popular approaches employ prompting strategies to elicit the alignment potential of MLLMs. As shown in Figure~\ref{fig:framework2}, this section discusses four prompting strategies for alignment without parameter tuning, which include (1) \textit{Direct Prompting}, (2) \textit{Code-Switching Prompting}, (3) \textit{Translation Alignment Prompting} and (4) \textit{Retrieval Augmented Alignment}.
\subsubsection{Direct Prompting} 
As shown in Figure~\ref{fig:framework2} (a), \textit{Direct Prompting} means directly outputting the request without any additional instruction for implicit alignment through MLLM itself~\cite{abdelali2023benchmarking,zhang2023cross,wang2023cross,wang2023document,lin2022few,bansal2023large,wei2023zero,pourkamali2024machine}.
\subsubsection{Code-Switching Prompting}
As shown in Figure~\ref{fig:framework2} (b), it integrates multilingual words into a single-language utterance, which is a typical language phenomenon~\cite{winata2022decades,dougruoz2023survey,dogruoz-etal-2023-representativeness} for effective language alignment~\cite{qin2020cosda,qin2022gl}.
Specifically, \citet{yong2023prompting,amin2023marathi} showed the effectiveness of MLLMs in cross-lingual alignment through model-generated code-switching texts. Furthermore, \citet{zhang2023multilingual} suggested the need for fairer and more detailed code-switching optimization for further research.

\subsubsection{Translation Alignment Prompting}
\textit{Translation alignment prompting} approaches mean that translating the query into other languages for better alignment (see Figure~\ref{fig:framework2} (c)), which can be divided into the following classes:
(1) \textit{Key Information Translation}: This approach focuses on extracting key information and executing translation for word-level cross-lingual alignment~\cite{lu2023chain,li2023bilingual}.
(2) \textit{Direct Translation}:  the model directly translates the whole input, enhancing alignment performance~\cite{etxaniz2023multilingual, zhang2023prompting,cheng2023scale,petrick2023document,hoang2023fly,zeng2023improving,nambi2023breaking,lin2021few}.
(3) \textit{Step-by-step Translation}: Instead of direct translation, this method prompts MLLMs to translate whole input step-by-step~\cite{puduppully2023decomposed,moslem2023adaptive,raunak2023leveraging,wu2023exploring,puduppully2023decomt,pilault2023interactive}.
(4) \textit{Restatement}: Beyond preserving original semantics, some studies focus on prompting MLLM to restate multilingual inputs to enhance cross-lingual effectiveness~\cite{shi2022language,patel2022bidirectional,rosenbaum2022clasp,asai2023buffet,qin2023crosslingual,huang2023not,tanwar2023multilingual}.
Further, considering the differences in multiple languages~\cite{ohmer2023evaluating}, \citet{qin2023crosslingual,ranaldi2023empowering} integrated knowledge and translation strategy across different languages by cross-lingual prompting.

\subsubsection{Retrieval Augmented Alignment}
\label{sec:dynamic_align}
\textit{Retrieval Augmented Alignment} incorporates external retrieval during prompting to inject more knowledge in MLLMs (see Figure~\ref{fig:framework2} (d)). Specifically, \citet{he2023exploring,zhang2023leveraging,conia2023increasing,xu2023language,ahmad2024enhancing} focus on retrieving cultural or professional knowledge to enrich prompts.
Another series of work focused on retrieval for high-quality alignment demonstrations, yielding significant improvements~\citep{shi2022xricl,agrawal2022context,li2023classification,winata2023multilingual,garcia2023unreasonable,li2023crosslingual,ramos2023lmcap,kim2023boosting,thakur2023nomiracl}.

\begin{takeaways}
\ \paragraph{Takeaways}
(1) \textit{Translation alignment prompting is more effective for cross-lingual alignment.}
(2) \textit{Retrieval augmented alignment mitigates knowledge gaps in LLM.}
\end{takeaways}

\section{Future work and New Froniter}\label{sec:future-work-new-froniter}

\subsection{Hallucination in MLLMs}
While remarkable progress has been achieved in MLLMs, the current approaches still face hallucination issues~\cite{raunak2021curious}.
Specifically, \citet{guerreiro2023hallucinations,aharoni2023multilingual,dale2023halomi,qiu2023detecting} have previously pointed out the hallucination phenomenon on current MLLM. Further, a series of works provide corresponding solutions in the pre-training~\cite{pfeiffer2023mmt5}, SFT~\cite{chen2023improving} and decoding~\cite{ahuja2022calibration,yang2023understanding,sia2023anti,zeng2023improving} stages.

The key challenges in this direction include:
(1) \textit{\textbf{Multilingual Hallucination Detection}}: 
How to effectively detect the hallucination phenomenon of MLLM across different languages is the primary problem to be solved in this field.
(2) \textit{\textbf{Multilingual Hallucination Alleviation}}: 
Current strategies for hallucination alleviation still focus on incorporating extensive factual data or utilizing external systems, which pose significant challenges for multiple languages, especially low-resource languages.

\subsection{Knowledge Editing in MLLMs}
The current MLLMs still face challenges with inaccurate, inconsistent, and outdated knowledge across different languages, which limits their performance.
To solve this issue, \citet{wu2023eva,wang2023cross-e} introduce a multilingual knowledge editing approach and propose a new benchmark for knowledge editing in MLLM. In addition, \citet{qi2023cross} introduce the cross-lingual consistency metric to ensure factual consistency across languages. Additionally, \citet{wang2023retrieval} incorporate a multilingual knowledge base into MLLMs with retrieval methods to facilitate knowledge editing.

The key challenges of this research include:
(1) \textit{\textbf{Continuous Knowledge Editing}}: How to continuously integrate new knowledge while preserving the accuracy of existing knowledge is a core challenge to explore.
(2) \textit{\textbf{Balancing Universal and Language-Specific Knowledge}}: Current work often neglects language-specific details like culture and slang, impacting user experience and causing cultural conflicts~\cite{held2023material,beniwal2024cross}.
How to balance universal knowledge, while preserving language-specific knowledge presents a fascinating question.

\subsection{Safety in MLLMs}
With the development and application of MLLMs, researchers have found that MLLMs often suffer some serious moral~\cite{costa2022toxicity,sanchez2023gender} and privacy~\cite{macko2023multitude} risks, hindering the development of MLLMs~\cite{wang2023all,ye2023multilingual,hammerl2022multilingual,shen2024language}.
Therefore, how to improve the safety of MLLMs is a promising research question.

The main challenges for safe MLLM are as follows:
\noindent (1) \textit{\textbf{Lack of Safety Benchmark}}: 
The lack of safe data in current literature hampers the relevant research. Consequently, acquiring a large-scale safety dataset to facilitate future research has become a hot topic.
\noindent (2) \textit{\textbf{Removal of Unsafe Data}}: 
The multilingual data generated by MLLMs poses potential unsafe risks during training~\cite{wang2023m4}. Therefore, identifying and filtering out unsafe multilingual content is a crucial issue~\cite{bogoychev2023opuscleaner}.

\subsection{Fairness in MLLMs}
Multilingual fairness refers to equal treatment and performance across languages and cultures~\cite{yu2022beyond,shliazhko2022mgpt}. But there is a significant performance gap between languages, especially on low-resource languages~\cite{malkin2022balanced,sengupta2023jais,ye2023language}.
Additionally, token consumption also varies by language in MLLMs, leading to unequal computational costs~\cite{koishekenov2022memory,hua2023lacos,nicosia2022evaluating,xue2022byt5,sun2023multi,rust2022language}.

The main concerns regarding fairness in MLLM are as follows:
(1) \textit{\textbf{Low-resource language performance improvement}}:
It is essential to improve the performance of low-resource languages with limited data~\cite{lin2023mplm,ansell2023unifying,adeyemi2023zero}.
(2) \textit{\textbf{Multilingual Token Cost Improvement}}: Current tokenizer exhibits biases in segmenting different languages, leading to varying token costs~\cite{petrov2023language,ahia2023all, ali2023tokenizer}. Addressing this challenge is essential for ensuring fairer tokenization across languages.

\subsection{Language Extension in MLLMs}

Due to the limited languages supported by current work, integrating new languages into existing MLLM is a promising direction to explore~\cite{kew2023turning,shaham2024multilingual}.
To this end, \citet{cui2023efficient,yang2023bigtrans} suggest adding languages through two-stage pre-training. \citet{yong2022bloom+} observe that adapter-based methods are more effective than continuous pre-training.

This challenge encompasses two main aspects: (1) \textit{\textbf{Multiple Languages Extension:}} How to dynamically and effectively extend the languages for MLLMs is an interesting research question. (2) \textit{\textbf{Original Languages Preserving:}}
Since the expansion of the model in other languages will harm the original language performance, how to prevent the language extension in MLLM from forgetting the previously learned language is a major challenge.

\subsection{Multi-Modality Extension in MLLMs}
Since the improvement in the usability of MLLM, a large amount of work has begun to further extend MLLM into visual modality~\cite{geigle2023mblip,chen2022pali,chen2023palix,chen2023pali3,ramos2023lmcap,bai2023qwen,zhou2023rc3,hu2023large,zhou2023accessible,he2023wanjuan,guo2023bridging}, speech modality~\cite{huang2023speech,huang2024multilingual,cheng2023mu}, video modality~\cite{team2023gemini} and even other modalities.

This field faces two main challenges: (1) \textit{\textbf{Complex Reasoning Exploration}}:  Current multi-modal MLLMs are limited to simple cross-modal cross-lingual tasks, with a need for more exploration in complex reasoning. (2) \textit{\textbf{Comprehensive Benchmark}}: The current literature lacks comprehensive benchmarks, which hinders progress and evaluation in this evolving field.

\section{Conclusion}
In this work, we present a comprehensive survey of the advancements in multilingual large language models (MLLMs). 
Specifically, we provide a new taxonomy for MLLMs from alignment perspectives, which can offer a unified view for researchers to understand the progress of MLLMs.
In addition, we highlight some emerging trends and frontiers as well as their corresponding challenges in MLLMs. We hope this work can facilitate the research and inspire more breakthroughs in MLLMs literature.

\bibliographystyle{acl_natbib}
\bibliography{anthology,custom}

\appendix
\clearpage

\begin{table*}[t]
	\centering
	\begin{adjustbox}{width=\textwidth}
		\begin{tabular}{lccccc}
			\toprule
			Dataset & Storage Size & Token Size & Language Size & Source & Latest Update Time
			\\
			\midrule
			\rowcolor{gray!8}\multicolumn{6}{c}{\textit{Manual}}\\
			\midrule
			Bible Corpus~\cite{mayer-cysouw-2014-creating} & 5.2G & - & 833 & - & May-2014 \\
			MultiUN~\cite{ziemski-etal-2016-united} & - & 0.3B & 7 & - & Dec-2014 \\
			IIT Bombay~\cite{kunchukuttan-etal-2018-iit} & - & 0.04B & 2 & - & Dec-2021 \\
			\midrule
			\rowcolor{gray!8}\multicolumn{6}{c}{\textit{Web Crawling}}\\
			\midrule
			CC-100~\cite{conneau2020unsupervised} & - & 208B & 116 & CommonCrawl & Oct-2022 \\
			mC4~\cite{xue2021mt5} & 38.5T & 6.3T & 101 & CommonCrawl & Oct-2022 \\
			Redpajamav2~\cite{together2023redpajama} & 30.4T & - & 5 & CommonCrawl & Dec-2023 \\
			OSCAR~\cite{suarez2019asynchronous} & 6.3T & 800B & 166 & CommonCrawl & Jan-2023 \\
			Oromo~\cite{ogueji2021small} & 0.939G & 0.1B & 11 & CommonCrawl & Feb-2022 \\
			Wu Dao 2.0 & - & 24B & 2 & CommonCrawl & Oct-2023 \\
			
			Europarl~\cite{koehn2005europarl} & 1.5G & 0.6B & 21 & - & May-2012 \\
			JW300~\cite{agic2019jw300} & - & 1.5B & 343 & - & Jul-2019 \\
			Glot500~\cite{imanigooghari2023glot500} & 600G & - & 511 & - & May-2023 \\
			Wikipedia~\cite{wikidump} & - & 24B & 300 & Wikipedia & - \\
			WikiMatrix~\cite{schwenk-etal-2021-wikimatrix} & 65G & - & 85 & Wikipedia & Apr-2021 \\
			OPUS-100~\cite{zhang-etal-2020-improving} & 2.6G & - & 100 & OPUS & Jul-2020 \\
			AfricanNews~\cite{adelani2022few} & 12.3G & - & 16 & mC4 & Sept-2023 \\
			
			Taxi1500~\cite{ma2023taxi1500} & - & - & 1500 & Bible Corpus & May-2023 \\
			CulturaX~\cite{nguyen2023culturax} & 27T & 6.3T & 167 & mC4, OSCAR & Jan-2024 \\
			
			\midrule
			\rowcolor{gray!8}\multicolumn{6}{c}{\textit{Benchmark Adaptation}}\\
			\midrule
			ROOTS~\cite{laurenccon2022bigscience}  & 1.6T & - & 46 & Huggingface & Jun-2022 \\
			OPUS~\cite{tiedemann-2012-parallel} & - & 40B & 1304 & - & Dec-2021 \\
			CCMT~\cite{yang2019ccmt}& - & - & 6 & - & - \\
			WMT~\cite{kocmi2023findings} & - & - & 32 & - & - \\
			IWSLT~\cite{agarwal2023findings} & 4.2G & - & 10 & - & - \\
			\bottomrule
		\end{tabular}
	\end{adjustbox}
	\caption{
		Pre-training Data Resource, where \colorbox{gray!8}{*} indicates different categories. The term ``Source'' refers to the origin datasets from which the pre-training data is derived.
	}
	\label{exp:pretrain-data}
\end{table*}

\section*{Appendix}
\section{Multilingual Performance Evaluation}
To facilitate the comparison of LLMs, extensive efforts have been invested in exploring enhanced evaluation methods for multilingual scenarios. This discussion will elaborate on MLLM evaluation, covering both (1) \textit{Evaluation Metrics} and (2) \textit{Evaluation Benchmarks}.
\subsection{Evaluation Metrics}

\paragraph{Traditional Automatic Metric} means that we assess predicted output using probabilities or pre-trained language model logits~\cite{liu2023revisiting,zouhar2024quality}.
Generally speaking, researchers use BLEU~\cite{papineni2002bleu}, BLEURT~\cite{sellam2020bleurt}, chrF++~\cite{popovic-2017-chrf} and COMET~\cite {rei2020comet} for translation evaluation,
and use ROUGE~\cite{lin-2004-rouge} for summary evaluation.
Further, \citet{guerreiro2023xcomet} proposed xCOMET for better translation evaluation through fine-grained error detection.
In assessing the general quality of generated text, the commonly employed approach is the utilization of multi-lingual BERTScore~\cite{Zhang2020BERTScore} as an evaluation metric.
\citet{qin2023crosslingual} extended Roscoe~\cite{Golovneva2022ROSCOEAS} to multi-language for quality assessment of multi-lingual CoT.
Further, \citet{hlavnova2023empowering} developed a comprehensive and robust multi-lingual checklist system to thoroughly assess the MLLMs' performance.

\paragraph{MLLM-based Automatic Metric} employs robust MLLMs to score or compare generated outputs for evaluation purposes~\cite{li2023bactrian, zhang2023bayling,vernikos2024don}. Specifically, \citet{zheng2023judging} introduced LLM-as-a-Judge, where GPT4 is prompted to assess the performance of other LLMs by comparing its output to the predicted one. However, this method discussed remains unreliable in multilingual settings~\cite{hada2023large}. And caution should be exercised, particularly in languages  where it is known that the MLLM has performed poorly. Furthermore, \citet{kim2023pr,muller2023evaluating} conducted attribution evaluation to deeply evaluate the robustness of the model.

\paragraph{Human Evaluation}
involves manually assessing MLLMs through detailed evaluation~\cite{zhang2023bayling,li2023bactrian,khondaker2023gptaraeval,zhang2023prompting}. 
\citet{lyu2023new} initially explored the multilingual challenges of ChatGPT through manually annotated cases.
Furthermore, \citet{hu2024dialight} introduced a new platform for more convenient manual assessments.

\subsection{Evaluation Benchmarks}
Current MLLMs tend to pay more attention to the alignment effect of non-English languages. Based on the different angles of alignment, we divide it into two categories: (1) \textit{Natural Language Understanding}; (2) \textit{Natural Language Generation}.
\begin{table*}[t]
	\centering
	\begin{adjustbox}{width=\textwidth}
		\begin{tabular}{lcccc}
			\toprule
			Dataset & Sample Size & Multi-lingual Instruction & Language Size & Task Size
			\\
			\midrule
			\rowcolor{gray!8}\multicolumn{5}{c}{\textit{Manual}}\\
			\midrule
			Sup-NatInst~\cite{wang2022super} & - & - & 55 & 1616 \\
			OpenAssist~\cite{kopf2023openassistant} & - & - & 35  & - \\
			EcomInstruct~\cite{li2023ecomgpt} & 2.5M & Yes & 2 & 12 \\
			COIG-PC-lite~\cite{COIG-PC} & 650k & No & 2 & 3,250 \\
			\midrule
			\rowcolor{gray!8}\multicolumn{5}{c}{\textit{Benchmark Adaption}}\\
			\midrule
			xP3~\cite{muennighoff2022crosslingual} & - & No & 71 & 46 \\
			BUFFET~\cite{asai2023buffet} & -  & - & 54 & 15 \\
			PolyglotPrompt~\cite{fu2022polyglot} & -  & No & 49 & 6 \\
			\midrule
			\rowcolor{gray!8}\multicolumn{5}{c}{\textit{Translation}}\\
			\midrule
			xP3-MT~\cite{muennighoff2022crosslingual} & -  & Yes & 46 & 71 \\
			MultilingualSIFT~\cite{Chen_MultilingualSIFT_Multilingual_Supervised_2023}  & -  & Yes & 11 & - \\
			Bactrian-X~\cite{li2023bactrian}  & -  & Yes & 52 & - \\
			MuIT~\cite{zhu2023extrapolating} & -  & Yes & 6 & - \\
			CrossAlpaca~\cite{ranaldi2023empoweringt} & - & - & 6 & - \\
			MGSM8KInstruct~\cite{chen2023breaking} & 73.6k & Yes & 6 & 10 \\
			XCoT~\cite{chai2024xcot} & 7.4K & Yes & 10 & 2 \\
			\midrule
			\rowcolor{gray!8}\multicolumn{5}{c}{\textit{MLLM Aided}}\\
			\midrule
			ShareGPT~\cite{sharegpt} & - & - & - & - \\
			Vicuna~\cite{vicuna2023} & - & - & - & - \\
			OverMiss~\cite{chen2023improving} & 54K & - & 3 & 1 (Translation) \\
			MultiAlpaca~\cite{wei2023polylm} & 133K & - & 11 & - \\
			Guanaco~\cite{dettmers2023qlora} & 535K & - & 5 & - \\
			Alpaca-4~\cite{peng2023instruction} & 52K & - & 2 & - \\
			\bottomrule
		\end{tabular}
	\end{adjustbox}
	\caption{
		Supervised Fine-Tuning Data Resource, where \colorbox{gray!8}{*} indicates different categories. The term ``Multi-lingual Instruction'' denotes the presence of instructions in multiple languages to form the specific data input.
	}
	\label{exp:sft-data}
\end{table*}

\label{sec:appendix-competency}
\subsubsection{Natural Language Understanding}
\paragraph{Linguistics Analysis}

For multilingual models, the most basic thing is to understand the linguistic differences between different languages~\cite{xu2023structural}. The most common multilingual linguistics assessment includes Part-of-Speech (POS)~\cite{liang2020xglue,11234/1-4758}, grammar analysis~\cite{kwon2023chatgpt,alhafni-etal-2023-advancements,michaelov2023structural,kwon2023beyond} and morphology~\cite{weissweiler-etal-2023-counting}.
Furthermore, \citet{zhang2023mela,song2022sling} conducted a comprehensive evaluation the linguistic acceptability of MLLM across languages.
\paragraph{Semantic Understanding}
Researchers take more care of should be able to analyze and understand the specific semantics of multiple languages~\cite{lai2023chatgpt,schott2023polyglot,panchendrarajan2024claim}. The most basic is to perform local semantic understanding, and the most typical one is the information extraction task~\cite {wei2023zero}, including: masakhaNER~\cite{adelani2021masakhaner}, MASSIVE~\cite{fitzgerald2022massive}, MultiCoNER~\cite{malmasi2022multiconer,fetahu2023multiconer}, WikiAnn~\cite{pan-etal-2019-cross} and SMiLER~\cite{seganti-etal-2021-multilingual}
The second is the semantic understanding of complete sentences, including: XNLI~\cite{conneau2018xnli}, Paws-X~\cite{yang2019paws}, MixATIS++~\cite{xu-etal-2020-end}, MTOP~\cite{li2020mtop}, MultiNLU~\cite{schuster2018cross}, and PRESTO~\cite{goel2023presto}.
Finally, there is the semantic understanding of the paragraph, like question-answering tasks with context: MLQA~\cite{lewis2019mlqa}, XQuAD\cite{artetxe-etal-2020-cross}, TyDiQA~\cite{clark2020tydi} and X-PARADE~\cite{rodriguez2023x}, X-CLAIM~\cite{mittal-etal-2023-lost}, Readme++~\cite{naous2023towards}, XKaggle-DBQA~\cite{shi2022xricl} and \citet{de-varda-marelli-2023-scaling}. 
Due to the emergence of a large number of multilingual benchmarks in recent years, a series of work has begun to combine the various existing semantic understanding tasks together for unified evaluation, including: XTREME~\cite{hu2020xtreme}, XTREME-R~\cite{ruder2021xtreme},  XGLUE~\cite{liang2020xglue}, MEGA~\cite{ahuja2023mega}, MEGAVerse~\cite{ahuja2023megaverse}, AGIEval~\cite{zhong2023agieval}, and Superlim~\cite{berdivcevskis2023superlim}.
Further, \citet{thapliyal2022crossmodal,changpinyo2022towards,fujinuma2023multi,kudugunta2023madlad} extend the semantic understanding of multi-modal context.
Since MLLMs have some biases~\cite{costa2023multilingual,lee2023target} or vulnerabilities~\cite{xu2023cognitive,puttaparthi2023comprehensive,shen2024language}, \citet{espana2023multilingual,cao2023multilingual,jiang2024cross,macko2024authorship} has begun to consider the corresponding benchmark to evaluate MLLMs.

\paragraph{Cultural Understanding}
Limited by cultural differences, the understanding between different languages is not completely parallel~\cite{li2023land,maity2023multilingual,cahyawijaya2023nusacrowd}, so researchers began to explore how to evaluate multi-cultural scenes~\cite{naous2023having,hershcovich2022challenges}. The most typical one is multi-cultural sentiment analysis. Sentiment Analysis~\cite{davidson2017automated,srinivasan2022tydip,li2023bactrian,muhammad2023afrisenti,winata-etal-2023-nusax,yadav2023lahm}.
Furthermore, \citet{zhang2023skipped} expands the multicultural scene to the entire Sociopragmatic Understanding level. In particular, \citet{kabra2023multi,wang2023cdeval,jiang2023cpopqa,fung2022normsage,li2023normdial,son2023hae,zhou2023red} proposed new benchmarks, requiring the model to fully understand different cultures. Furthermore, due to the emergence of reasoning capabilities, \citet{qin2023crosslingual,liu2023multilingual,wang2023seaeval} start to evaluate the reasoning ability of MLLMs with different cultural backgrounds.

\paragraph{Knowledge Understanding}
A large amount of work has been done to test the degree of knowledge transfer of MLLM between different languages through examination questions.
Specifically, \citet{hardalov2020exams,xuan2023vnhsge,zhang2023evaluating} proposed for the comprehensive knowledge test at the high school level in a multilingual scenario. \citet{zhang2023don} design a complex translation strategy to translate existing benchmark for multilingual evaluation. On this basis, M3Exam~\cite{zhang2023m3exam} further expands comprehensive testing to multi-language and multi-modal scenarios.
Furthermore, \citet{gekhman2023trueteacher} tested the factual consistency of MLLM. And \citet{choudhury2023ask,joseph2023multilingual,zhao2023chatagri,wei2023zero,goenaga2023explanatory,datta2023mildsum,thulke2024climategpt} proposed benchmarks to evaluate the multilingual  scientific and applied professional domain knowledge for current MLLMs.

\subsubsection{Natural Language Generation}

\paragraph{Translation}
In the process of multi-lingual alignment, in addition to testing whether the multiple languages are aligned in terms of understanding capabilities, researchers often also need to consider whether the two can be aligned in terms of output capabilities. The most typical task is machine translation~\cite{dabre2020survey,vilar2022prompting}, currently commonly used data sets are: FLORES-101~\cite{goyal2022flores}, FLORES-200~\cite{costa2022no}, WMT~\cite{kocmi2023findings} and DiaBLa~\cite{bawden2021diabla}.
Recently, \citet{lou2023cceval} proposed CCEval for Chinese-centric translation for comprehensive evaluation on MLLMs.
Furthermore, due to the large gap between languages~\cite{zhang2023don,choudhury2023ask,mujadia2023assessing,fujii2023different,khatri2023study,etxaniz2023multilingual,artetxe-etal-2023-revisiting,stefanovitch2023holistic,ramesh2023fairness,deng2022knowledge,held2023material,bang2023multitask,lai2023chatgpt,philippy2023towards,ye2023language}, \citet{kuparinen2023dialect,wassie2023machine,liu2023nlebench+,rakhimova2024task} focused more on low-resource language translation. Additionally, \citet{yang2023investigating,gueuwou2023jwsign,bellagente2023multifusion,zhong2023let,tuo2023anytext} further extend the translation and restatement tasks into multi-modal settings for practical scenario.

\paragraph{Reasoning}
Currently, the most commonly used reasoning ability assessments of MLLMs tend to focus on commonsense and mathematical reasoning~\cite{huang2023not,qin2023crosslingual}. Specifically, commonsense reasoning includes XCOPA~\cite{ponti2020xcopa}, MARC~\cite{keung2020multilingual}, XWinograd~\cite{tikhonov2021s}, GEOMLAMA~\cite{yin2022geomlama}, X-CSQA~\cite{lin-etal-2021-common}, XStoryCloze~\cite{lin-etal-2022-shot}, ASPEN~\cite{razumovskaia2022little} and Masakhanews~\cite{adelani2023masakhanews} . Additionally, mathematical reasoning includes MGSM~\cite{shi2022language} and WizardMath~\cite{luo2023wizardmath}. Due to the expensive annotations for multilingual reasoning, \citet{zhang2023don} propose a complex translation and filter process to construct a multilingual reasoning benchmark.

\paragraph{Coding Generation}
Coding generation requires that MLLMs can generate structured, executable code programs. The common-used benchmarks include XSPIDER~\cite{shi2022xricl}, XSEMPLR~\cite{zhang2023xsemplr}, ODEX~\cite{wang2022execution} and Mconala~\cite{wang2022mconala}.
\paragraph{Summarization}
To test the summarization ability of the model, the model is required to be able to summarize key information based on long texts. The simplest one, \citet{ryan2023revisiting} proposed a multi-lingual text reduction benchmark for the evaluation of MLLM. Secondly, a lot of work focuses on cross-lingual summarization. Typical data sets include: XSUM~\cite{narayan2018don}, and CrossSum~\cite{bhattacharjee2021crosssum}.
On this basis, \citet{wang2022clidsum} introduced multilingual conversation summarization, and \citet{zhang2023crocosum} proposed the concept of code-switch in the evaluation, making it more practical. \citet{urlana2023pmindiasum} further proposed headline summarization for languages in India. SEAHORSE~\cite{clark2023seahorse} further extended them to the multifaceted multilingual summarization. In addition \citet{nguyen2023loralay,verma2023large} developed summarization benchmarks for multi-modal scenarios.
\paragraph{Dialogue}
The communication between models and humans is often interactive, so a lot of work pays attention to MLLMs' dialogue ability~\cite{boughorbel2023analyzing}. The current evaluation set includes xDial-Eval~\cite{zhang2023xdial}, Multi$^3$WOZ~\cite{hu2023multi3woz}, DIALIGHT~\cite{hu2024dialight}, HPD~\cite{chen2023large} and X-RiSAWOZ~\cite{moradshahi2023x}. Since multiple rounds of dialogue are not controllable, traditional indicators cannot be used. Currently, we tend to use PLM for evaluation~\cite{mendoncca2023towards}. Furthermore, \citet{mendoncca2023simple} proposed a new benchmark, which can achieve more robust evaluation by coordinating with pretrained language models.
\citet{ferron2023meep} proposed the MEEP benchmark to further evaluate the dialogue participation of MLLMs.

\end{document}